\documentclass{llncs}
\usepackage{hyperref}
\usepackage{amsfonts}
\usepackage{amsmath}
\usepackage{booktabs} 
\usepackage{multirow}
\usepackage{graphicx}
\usepackage{float}
\usepackage{subcaption}
\usepackage{breakcites}
\usepackage{booktabs,colortbl,tabularx}

\begin{document}
\setlength{\abovedisplayskip}{0pt}
\setlength{\belowdisplayskip}{0pt}
\setlength{\abovedisplayshortskip}{0pt}
\setlength{\belowdisplayshortskip}{0pt}
\title{Question Dependent Recurrent Entity Network for Question Answering}
\titlerunning{QDREN}  
%
\author{Andrea Madotto\inst{1} \and Giuseppe Attardi\inst{1}}
\authorrunning{Madotto et al.} 
%
\tocauthor{Andrea Madotto, Giuseppe Attardi}
\institute{Dipartimento di Informatica, University of Pisa\\
  Largo B. Pontecorvo, 3\\
\email{andreamad8@gmail.com}, \email{attardi@di.unipi.it}
}

\maketitle              

\begin{abstract}
Question Answering is a task which requires building models capable of providing answers to questions expressed in human language. Full question answering involves some form of reasoning ability. We introduce a neural network architecture for this task, which is a form of \textit{Memory Network}, that recognizes entities and their relations to answers through a focus attention mechanism. Our model is named \emph{Question Dependent Recurrent Entity Network} and extends \emph{Recurrent Entity Network} by exploiting aspects of the question during the memorization process. We validate the model on both synthetic and real datasets: the \emph{bAbI} question answering dataset and the \emph{CNN \& Daily News} \emph{reading comprehension} dataset. In our experiments, the models achieved a State-of-The-Art in the former and competitive results in the latter.
\end{abstract}
\section{Introduction} 
Question Answering is a task that requires capabilities beyond simple NLP since it involves both linguistic techniques and inference abilities. Both the document sources and the questions are expressed in natural language, which is ambiguous and complex to understand. To perform such a task, a model needs in fact to understand the underlying meaning of text. Achieving this ability is quite challenging for a machine since it requires a reasoning phase (chaining facts, basic deductions, etc.) over knowledge extracted from the plain input data. In this article, we focus on two Question Answering tasks: a \emph{Reasoning Question Answering} (RQA) and a \emph{Reading Comprehension} (RC). These tasks are tested by submitting questions to be answered directly after reading a piece of text (e.g. a document or a paragraph). 

Recent progress in the field has been possible thanks to machine learning algorithms which automatically learn from large collections of data. Deep Learning \cite{lecun2015deep} algorithms achieve the current State-of-The-Art in our tasks of interest. A particularly promising approach is based on \emph{Memory Augmented Neural Networks}. These networks are also known as \emph{Memory Networks} \cite{weston2014memory} or \emph{Neural Turing Machines} \cite{graves2014neural}. In the literature the RQA and RC tasks are typically solved by different models. However, the two tasks share a similar scope and structure. We propose to tackle both with a model called \emph{Question Dependent Recurrent Entity Network}, which improves over the model called \emph{Recurrent Entity Network} \cite{henaff2016tracking}.

Our major contributions are: 1) exploiting knowledge of the question for storing relevant facts in memory, 2) adding a tighter regularisation scheme, and 3) changing the activation functions. We test and compare our model on two datasets, bAbI~\cite{weston2015towards} and \cite{hermann2015teaching}, which are standard benchmark for both tasks. The rest of the paper is organised as follows: section \textit{Related} outlines the models used in QA tasks, while section \textit{Model} the proposed QDREN model. Section \textit{Experiments and Results} show training details and performance achieved by our model. The section \textit{Analysis} reports a visualisation with the aim to explain the obtained results. Finally, section \textit{Conclusions} summarise the work done.

\section{Related Work}

\subsection{Reasoning Question Answering}
A set of synthetic tasks, called bAbI \cite{weston2015towards}, has been proposed for testing the ability of a machine in chaining facts, performing simple inductions or deductions. The dataset is available in two sizes, 1K and 10K training samples, and in two  settings, i.e. with and without supporting facts. The latter allows knowing which facts in the input are needed for answering the question (i.e. a stronger supervision). Obviously, the 1K sample setting with no supporting facts is quite hard to handle, and it is still an open research problem. \emph{Memory Network} \cite{weston2014memory} was one of the first models to provide the ability to explicitly store facts in memory, achieving good results on the bAbI dataset. An evolution of this model is the \emph{End to End Memory Network} \cite{sukhbaatar2015end}, which allows for end-to-end training. This model represents the State-of-The-Art in the bAbI task with 1K training samples. Several other models have been tested in the bAbI tasks achieving competitive results, such as \emph{Neural Turing Machine} \cite{graves2014neural}, \emph{Differentiable Neural Computer} \cite{graves2016hybrid} and \emph{Dynamic Memory Network} \cite{kumar2015ask,xiong2016dynamic}. Several other baselines have also been proposed \cite{weston2015towards}, such as: an $n$-gram \cite{richardson2013mctest} models, an LSTM reader and an SVM model. However, some of them still required strong supervision by means of the supporting facts. 

\subsection{Reading Comprehension}
\emph{Reading Comprehension} is defined as the ability to read some text, process it, and understand its meaning. A impending issue for tackling this task was to find suitably large datasets with human annotated samples. This shortcoming has been addressed by collecting documents which contain easy recognizable short summary, e.g. news articles, which contain a number of bullet points, summarizing aspects of the information contained in the article. Each of these short summaries is turned into a fill-in question template, by selecting an entity and replacing it with an anonymized placeholder.

Three datasets follows this style of annotation: \emph{Children’s Text Books} \cite{hill2015goldilocks}, \emph{CNN \& Daily Mail news articles} \cite{hermann2015teaching}, and \emph{Who did What} \cite{onishi2016did}. It is also worth to mention \emph{Squad} \cite{rajpurkar2016squad}, a human annotated dataset from Stanford NLP group. \emph{Memory Networks}, described in the previous sub-section, has been tested \cite{hill2015goldilocks} on both the CNN and CBT datasets, achieving good results. The \emph{Attentive and Impatient Reader} \cite{hermann2015teaching} was the first model proposed for the \emph{CNN and Daily Mail} dataset, and it is therefore often used as a baseline. While this model achieved good initial results, shortly later a small variation to such model, called \emph{Standford Attentive Reader} \cite{chen2016thorough}, increased its accuracy by 10\%. Another group of  models are based on an Artificial Neural Network architecture called \emph{Pointer Network} \cite{vinyals2015pointer}. \emph{Attentive Sum Reader} \cite{kadlec2016text} and \emph{Attention over Attention} \cite{cui2016attention} use a similar idea for solving different reading comprehension tasks. \emph{EpiReader} \cite{trischler2016natural} and \emph{Dynamic Entity Representation} \cite{kobayashi2016dynamic}, partially follow the \emph{Pointer Network} framework but they also achieve impressive results in the RC tasks. Also for this task several baselines, both learning and non-learning, have been proposed. The most commonly used are: \emph{Frame-Semantics}, Word distance, and \emph{LSTM Reader} \cite{hermann2015teaching} and its variation (windowing etc.).

\section{Proposed Model}
Our model is based on the \emph{Recurrent Entity Network} (REN)~\cite{henaff2016tracking} model. The latter is the only model able to pass all the 20 bAbI tasks using the 10K sample size and without any supporting facts. However, this model fails many tasks with the 1K setting, and it has not been tried on more challenging RC datasets, like the CNN news articles. Thus, we propose a variant to the original model called \emph{Question Dependent Recurrent Entity Network} ($QDREN$)\footnote{An implementation is available at \href{https://github.com/andreamad8/QDREN}{https://github.com/andreamad8/QDREN}}. This model tries to overcome the limitations of the previous approach. The model consists in three main components: \emph{Input Encoder}, \emph{Dynamic Memory}, and \emph{Output Module}.

The training data consists of tuples $\{(x_i,y_i)\}_{i=1}^n$, with $n$ equal to the sample size, where: $x_i$ is composed by a tuple $(T, q)$, where $T$ is a set of sentences $\{ s_{1},\dots,s_{t}\}$, each of which has $m$ words, and $q$ a single sentence with $k$ words representing the question. Instead, $y_i$ is a single word that represents the answer. \\
\textit{The Input Encoder} transforms the set of words of a sentence $s_{t}$ and the question $q$ into a single vector representation by using a multiplicative mask. Let's define $E\in \mathbb{R}^{|V|\times d}$ the embedding matrix, that is used to convert words to vectors, i.e. $E(w)=e \in \mathbb{R}^d$. Hence, $\{e_{1},\dots,e_{m} \}$ are the word embedding of each word in the sentence $s_{t}$ and $\{e_{1},\dots,e_{k}\}$ the embedding of the question's words. The multiplicative masks for the sentences are defined as $f^{(s)} = \{ f_1^{(s)},\dots,f_m^{(s)}\}$ and $f^{(q)} = \{ f_1^{(q)},\dots,f_m^{(q)}\}$ for the question, where each $f_i \in \mathbb{R}^d$. The encoded vector of a sentence is defined as:
\begin{equation}
s_{t} = \sum_{r=1}^m e_{r} \odot f_r^{(s)} \qquad \qquad q= \sum_{r=1}^k e_{r} \odot f_r^{(q)}  \nonumber
\end{equation} 
\textit{Dynamic Memory} stores information of entities present in $T$. This module is very similar to a Gated Recurrent Unit (GRU)~\cite{GRU} with a hidden state divided into blocks. Moreover, each block ideally represents an entity (i.e. person, location etc.), and it stores relevant facts about it. Each block $i$ is made of a hidden state $h_i\in \mathbb{R}^d$ and a key $k_i\in \mathbb{R}^d$, where $d$ is the embedding size. The Dynamic Memory module is made of a set of blocks, which can be represent with a set of hidden states $\{ h_1,\dots,h_z \}$ and their correspondent set of keys $\{ k_1,\dots,k_z \}$. The equation used to update a generic block $i$ are the following: 

\begin{align*}
g_i^{(t)} =& \sigma(s_t^T h_i^{(t-1)} + s_t^T k_i^{(t-1)} + s_t^T q ) &\text{(Gating Function)}&\\  
\hat h_i^{(t)} =& \phi(U h_i^{(t-1)} + V k_i^{(t-1)} + W s_t ) &\text{(Candidate Memory)}&\\ 
h_i^{(t)} =& h_i^{(t-1)} + g_i^{(t)} \odot \hat h_i^{(t)}  &\text{(New Memory)}&\\
h_i^{(t)} =& h_i^{(t)}/\| h_i^{(t)} \|  &\text{(Reset Memory)}&\\
\end{align*}
where $\sigma$ represents the sigmoid function, $\phi$ a generic activation function which can be chosen among a set (e.g. sigmoid, ReLU, etc.). $g_i^{(t)}$ is the gating function which determines how much the $i$th memory should be updated, and $\hat h_i^{(t)}$ is the new candidate value of the memory to be combined with the existing one $h_i^{(t-1)}$. The matrix $U \in \mathbb{R}^{d \times d}$, $V \in \mathbb{R}^{d \times d}$, $W \in \mathbb{R}^{d \times d}$ are \textbf{shared} among different blocks, and are trained together with the key vectors. 
\begin{figure}[t]
\centering
\includegraphics[scale=0.35]{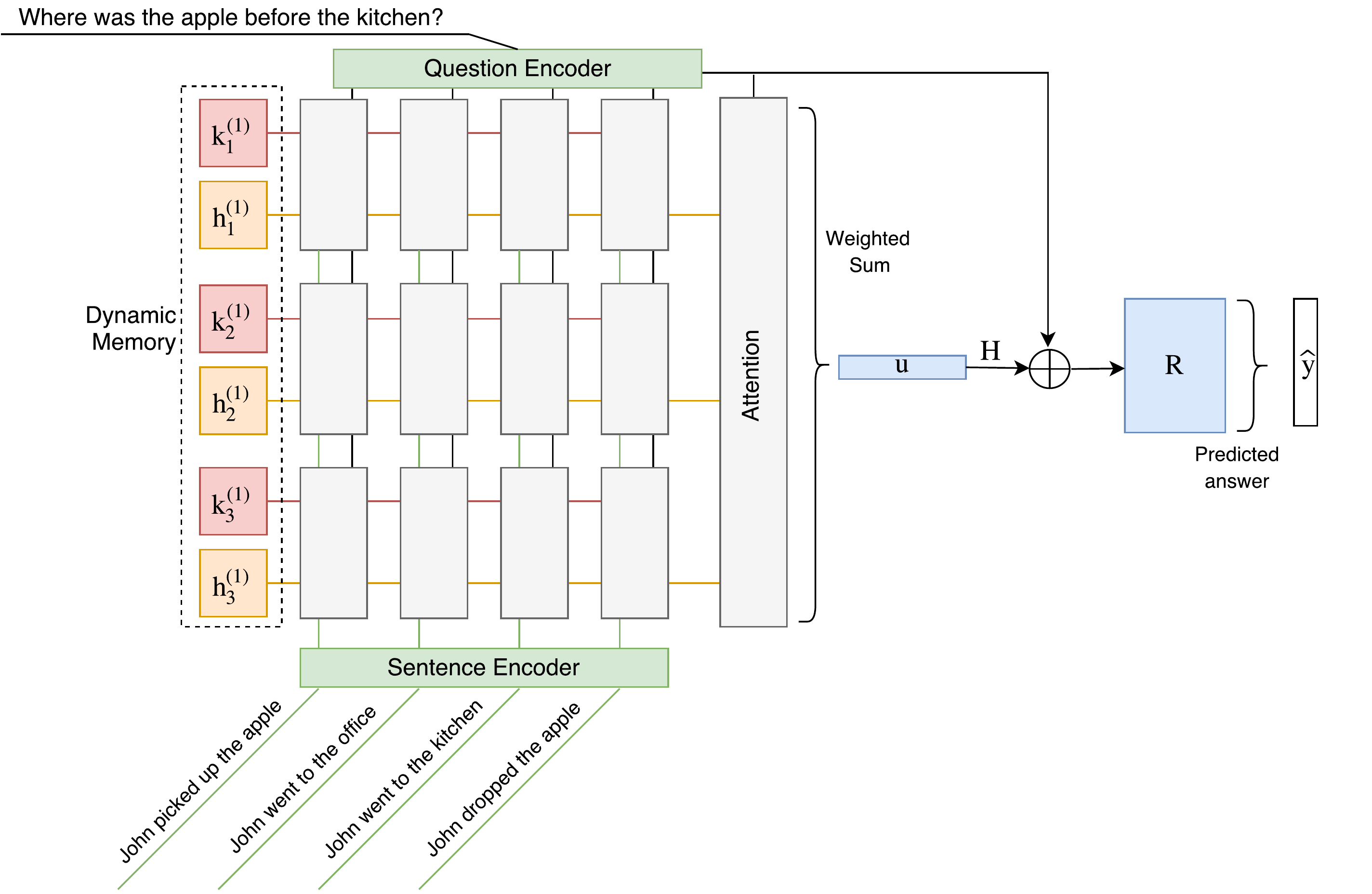}
\caption{Conceptual schema of the QDREN model, with three memory blocks. In input a sample taken from bAbI task dataset.}\label{QDREN}
\end{figure}
The addition of the $s_t^T q$ term in the gating function is our main contribution. We add such term with the assumption that the question can be useful to focus the attention of the model while analyzing the input sentences.
\\
\\
\textit{The Output Module} creates a probability distribution over the memories' hidden states using the question $q$. Thus, the hidden states are summed up, using the probability as weight, to obtain a single state representing all the input. Finally, the network output is obtained by combining the final state with the question. Let us define $R \in \mathbb{R}^{|V|\times d }$, $H \in \mathbb{R}^{d \times d}$, $\hat y \in \mathbb{R}^{|V|}$, $z$ is the number of blocks, and $\phi$ can be chosen among different activation functions. Then, we have:\\
\begin{align*}
p_i =& \text{Softmax}(q^T h_i)\\  
u =& \sum_{j=1}^{z} p_j h_j \\ 
\hat y =& R \phi(q + H u)
\end{align*}
The model is trained using a cross entropy loss $H(\hat y, y)$ plus L2 regularisation term, where $y$ is the one hot encoding of the correct answer. The sigmoid function and the L2 term are two novelty added to the original REN. Overall, the trainable parameters are:
\begin{equation}
\Theta = [E,f^{(s)},f^{(q)}, U, V, W, k_1,\dots,k_z, R, H ] \nonumber
\end{equation}
where $f^{(s)}$ refers to the sentence multiplicative masks, $f^{(q)}$ to the question multiplicative masks, and each $k_i$ to the key of a generic block $i$. The number of parameters is dominated by $E$ and $R$, since they depend on the vocabulary size. However, $R$ is normally is much smaller than $E$ like in the CNN dataset, in which the prediction is made on a restricted number of entities\footnote{Therefore $R \in \mathbb{R}^{|entities|\times d }$}. 
All the parameters are learned using the Backpropagation Through Time (BPTT) algorithm. A schematic representation of the model is shown in Figure~\ref{QDREN}.

\section{Experiments and Results}
Our model has been implemented using TensorFlow v1.1~\cite{tensorflow2015-whitepaper} and the experiments have been run on a Linux server with 4 Nvidia P100 GPUs. As mentioned earlier, we tested our model in two datasets: the bAbI 1k sample and the CNN news articles. The first dataset have 20 separate tasks, each of which has 900/100/1000 training, validation, and test samples. Instead, the second one has 380298/3924/3198 training, validation and test samples. We kept the original splitting to compare our results with the existing ones. 

\paragraph{bAbI:} in these tasks, we fixed the batch size to 32, we did not use any pre-trained word embedding, and we used Adam \cite{adam} optimizer. We have also clipped the gradient to a maximum of 40 (to avoid gradient explosion), and we set the word embedding size to 100, as it has also been suggested in the original paper. We have also implemented an early stopping method, which stop the training ones the validation accuracy does not improve after 50 epochs. Several values for the hyper-parameter have been tried and, for each task, we selected the setting that achieved the highest accuracy in validation. Once we selected the best model, we estimate its generalization error using the provided Test set. Table~\ref{hyper-babi} shows an example of the dataset and the used hyper-parameters.
\begin{table}[h]
    \caption{On the left an example of the bAbI task, and on the right the selected model hyper-parameters.\\}
          \label{hyper-babi}
\scalebox{0.82}{

    \begin{minipage}{.63\linewidth}
      \centering
      \begin{tabular}{l|l}
      \hline
      \textbf{Story} & \textbf{Question} \\ \hline
      \multirow{5}{*}{\begin{tabular}[c]{@{}l@{}}John picked up the apple\\ John went to the office\\ John went to the kitchen\\ John dropped the apple\end{tabular}} & \multirow{3}{*}{\begin{tabular}[c]{@{}l@{}}Where was the apple \\ before the kitchen?\end{tabular}} \\
       &  \\
       &  \\ \cline{2-2} 
       & \textbf{Answer} \\ \cline{2-2} 
       & office \\ \hline
      \end{tabular}

    \end{minipage}%
    \hfill
    \begin{minipage}{.63\linewidth}
    \centering
      \begin{tabular}{cc}
      \toprule
      \textbf{Parameter} & \textbf{Values}   \\ \midrule
      Learning Rate ($\alpha$)     & 0.01,0.001,0.0001 \\
      Number of Blocks 			   & 20,30,40,50       \\
      L2 reg. ($\lambda$)          & 0,0.001,0.0001    \\
      Dropout (\textbf{Dr})            & 0.3,0.5,0.7        \\ \bottomrule
      \end{tabular}
    \end{minipage} 
}
\end{table}
We compared our results with four models: $n$-gram model, LSTM, original REN (with no question in the gating function) and \emph{End To End Memory Network} (MemN2N) \cite{sukhbaatar2015end}, which is currently the State-Of-The-Art in this setting. To the best of our knowledge we achieved the lowest number of failed tasks, failing just 8 tasks compared with the previous State-Of-The-Art which was 11. Comparing our QDREN with the original \emph{Recurrent Entity Network} (REN) we achieved, on average, an improvement of 11\% in the average error rate and we passed 7 tasks more. Table~\ref{babi-comparisons} shows the error rate\footnote{The error is the percentage of wrong answers.} in the test set obtained using each compared model, and the hyper-parameter setting used in each task. We improve the mean error compared to the original REN, however we still do know reach the error rate achieved by the \emph{End To End Memory Network} (even if we passed more tasks). It is worth to notice the following two facts: first, in task 14 and 18 the error is very close to the threshold for passing the task (5\%); second, in task 2, we achieved a slightly worse result (10\% error more) with respect to the original REN. 

\begin{table}[t]
\centering
\caption{Test set error rate comparison between n-gram, LSTM, QDREN, REN and End To End Memory Network (MemN2N). All the results have been taken from the original articles. In bold we highlight the task in which we greatly outperform the other models. On the right the hyper-parameters used in QDREN. \\}
\label{babi-comparisons}
\scalebox{0.7}{
\begin{tabular}{c|ccccc|cccc}
\hline
\textbf{Task} & \textbf{$n$-gram} & \textbf{LSTM} & \textbf{MemN2N} & \textbf{REN} & \textbf{QDREN} & \textbf{Blk} & \textbf{$\lambda$} & \textbf{$\alpha$} & \textbf{Dr} \\ \hline
1 & 64 & 50 & 0 & 0.7 & 0 & 20 & 0 & 0.001 & 0.5 \\
2 & 98 & 80 & 8.3 & 56.4 & 67.6 & 30 & 0 & 0.001 & 0.5 \\
3 & 93 & 80 & 40.3 & 69.7 & 60.8 & 40 & 0 & 0.001 & 0.5 \\
4 & 50 & 39 & 2.8 & 1.4 & 0 & 20 & 0 & 0.001 & 0.5 \\
5 & 80 & 30 & 13.1 & 4.6 & \textbf{2.0} & 50 & 0 & 0.001 & 0.2 \\
6 & 51 & 52 & 7.6 & 30 & 29 & 30 & 0 & 0.001 & 0.5 \\
7 & 48 & 51 & 17.3 & 22.3 & \textbf{0.7} & 30 & 0 & 0.001 & 0.5 \\
8 & 60 & 55 & 10 & 19.2 & \textbf{2.5} & 20 & 0.001 & 0.001 & 0.7 \\
9 & 38 & 36 & 13.2 & 31.5 & \textbf{4.8} & 40 & 0.0001 & 0.001 & 0.5 \\
10 & 55 & 56 & 15.1 & 15.6 & \textbf{3.8} & 20 & 0 & 0.001 & 0.5 \\
11 & 71 & 28 & 0.9 & 8 & \textbf{0.6} & 20 & 0 & 0.001 & 0.5 \\
12 & 91 & 26 & 0.2 & 0.8 & 0 & 20 & 0 & 0.0001 & 0.5 \\
13 & 74 & 6 & 0.4 & 9 & \textbf{0.0} & 40 & 0.001 & 0.001 & 0.7 \\
14 & 81 & 73 & 1.7 & 62.9 & 15.8 & 30 & 0.0001 & 0.001 & 0.5 \\
15 & 80 & 79 & 0 & 57.8 & \textbf{0.3} & 20 & 0 & 0.001 & 0.5 \\
16 & 57 & 77 & 1.3 & 53.2 & 52 & 20 & 0.001 & 0.001 & 0.5 \\
17 & 54 & 49 & 51 & 46.4 & 37.4 & 40 & 0.001 & 0.001 & 0.5 \\
18 & 48 & 48 & 11.1 & 8.8 & 10.1 & 30 & 0.0001 & 0.001 & 0.5 \\
19 & 10 & 92 & 82.8 & 90.4 & 85 & 20 & 0 & 0.001 & 0.5 \\
20 & 24 & 9 & 0 & 2.6 & 0.2 & 20 & 0 & 0.001 & 0.5 \\ \hline
\multicolumn{1}{l|}{Failed Tasks (\textgreater5\%):} & 20 & 20 & 11 & 15 & 8 &  &  &  &  \\
\multicolumn{1}{l|}{Mean Error:} & 65.9 & 50.8 & 13.9 & 29.6 & 18.6 &  &  &  &  \\ \hline
\end{tabular}
}
\end{table}

\paragraph{CNN news articles:} in this dataset, the entities in the original paragraph are replaced by an ID, making the task even more challenging. The CNN dataset is already tokenized and cleaned, therefore we did not apply any text pre-processing. As it was done in other models, the set of possible answers is restricted to the set of hidden entities in the text, that are much less, around 500, compared to all the words (120K) in the vocabulary. Compared to the model used for bAbI, we changed the activation function of the output layer, using a sigmoid instead of parametric ReLU, since after several experiments we noticed that such activation was hurting the model performance. Moreover, the input was not split into sentences, thus we divided the text into sentences using the dot token ("."). sentence splitting in general is itself a challenging task, but in this case the input was already cleaned and normalised. However, the sentence may be very long, thus we intrdocued a windowing mechanism. The same approach has been used in the \emph{End To End Memory Network} \cite{sukhbaatar2015end} as a way to encode the input sentence. This method takes each entity marker ($@entity_i$) and it creates a window of $b$ words around it. Formally, $\{ w_{i-\frac{(b-1)}{2}}, \dots, w_i, \dots,w_{i+\frac{(b-1)}{2}} \}$, where $w_i$ represent the entity of interest. For the question, a single window is created around the placeholder marker (the word to predict). Moreover, we add $2(b-1)$ tokens for the entities at the beginning and at the end of the text. To check whether our QDREN could improve the existent REN and whether the window-based approach makes any difference in comparison with plain sentences, we separately trained four different models: REN+SENT, REN+WIND, QDREN+SENT and QDREN+WIND. Where SENT represent simple input sentences, and WIND the window as a input. For each of this model, we conduct a separated model selection using a various number of hyper-parameters. Table~\ref{hyper-CNN} shows an example of the dataset and the used hyper-parameters.
\begin{table}[h]
    \caption{On the left, an example from CNN news article, and on the right, the model selection Hyper-parameters.}
          \label{hyper-CNN}
\scalebox{0.8}{

    \begin{minipage}{.63\linewidth}
      \centering
       \begin{tabular}{l|l}
      \hline
      \textbf{Story} & \textbf{Question} \\ \hline
      \multirow{5}{*}{\begin{tabular}[c]{@{}l@{}}( @entity1 ) @entity0 may \\ be @entity2 in the popular \\ @entity4 superhero films \\ but he recently dealt in some \\ advanced bionic technology ... \end{tabular}} & \multirow{3}{*}{\begin{tabular}[c]{@{}l@{}}"@placeholder" star \\ @entity0 presents a \\ young child \end{tabular}} \\
       &  \\
       &  \\ \cline{2-2} 
       & \textbf{Answer} \\ \cline{2-2} 
       & @entity2 \\ \hline
      \end{tabular}
    \end{minipage}%
    \hfill
    \begin{minipage}{.63\linewidth}
    \centering
    \begin{tabular}{cc}
      \toprule
    \textbf{Parameter} & \textbf{Values} \\ \hline
    Learning Rate ($\alpha$) & 0.1,0.01,0.001,0.0001 \\
    Window & 2,3,4,5 \\
    Number of Blocks & 10,20,50,70,90 \\
    L2 reg. ($\lambda$) & 0.0,0.001,0.0001,0.00001 \\
    Optimizer & Adam,RMSProp \\
    Batch Size & 128,64,32 \\
    Dropout (\textbf{Dr})  & 0.2,0.5,0.7,0.9 \\ \hline
    \end{tabular}
    \end{minipage} 
}
\end{table}
As for the bAbI task, we used early stopping, ending the training once the validation accuracy does not improve for 20 epochs. Since each training required a large amounts of time (using a batch size of 64 an epoch takes around 7 hours), we opted for a random search technique~\cite{RandomSearch}, and we used just a sub-sample of the training set, i.e. 10K sample, for the model selection, but we still keep the validation set as it was. Obviously, this is not an optimal parameter tuning, since the model is selected on just 10K samples. Indeed, we noticed that the selected model, which is trained using all the samples (380K), tends to under-fit. However, it was the only way to try different parameters in a reasonable amount of time. Moreover, we also limited the vocabulary size to the most common 50K words, and we initialize the embedding matrix using Glove \cite{pennington2014glove} pre-trained word embedding of size 100. 
\begin{table}[h]
\centering
\caption{Test set accuracy comparison between \textbf{REN+SENT}, \textbf{QDREN+SENT}, \textbf{REN+WIND} and \textbf{QDREN+WIND}. We show the best hyper-parameters picked by the model selection, and the accuracy values.\\}
\label{CNN-ris}
\scalebox{0.7}{
\begin{tabular}{@{}r|cccc@{}}
\toprule
\multicolumn{1}{l|}{} & \textbf{REN+SENT} & \textbf{QDREN+SENT} & \textbf{REN+WIND} & \textbf{QDREN+WIND} \\ \midrule
Number of Blocks    & 20     & 10    & 50     & 20      \\ 
Window              & -      & -     & 5      & 4       \\ 
Learning Rate       & 0.001  & 0.001 & 0.0001   & 0.01    \\ 
Optimizer           & Adam   & Adam  & RMSProp   & RMSProp \\ 
Dropout             & 0.7    & 0.2   & 0.5    & 0.5     \\ 
Batch Size          & 128    & 64    & 64     & 64      \\ 
$\lambda$           & 0.0001 & 0.001 & 0.0001 & 0.0001 \\ \midrule
Loss Training       & 2.235  & 2.682 & 2.598  & 2.216   \\
Loss Validation     & 2.204  & 2.481 & 2.427  & 1.885   \\
Loss Test           & 2.135  & 2.417 & 2.319  & 1.724   \\ \midrule
Accuracy Training   & 0.418  & 0.349 & 0.348  & 0.499   \\
Accuracy Validation & 0.420  & 0.399 & 0.380  & 0.591   \\
Accuracy Test       & 0.420  & 0.397 & 0.401  & \textbf{0.628}  \\ \bottomrule
\end{tabular}
}
\end{table}
As before, we selected the models that achieved the highest accuracy in the validation set, and then we estimate its generalization error using the provided test set. The selected models, with their hyper-parameters, are shown in Table~\ref{CNN-ris}. The best accuracy\footnote{Percentage of correct answers.} is achieved by QDREN+WIND with a value of 0.628, while all other models could not achieve an accuracy greater than 0.42. The window-based version without question supervision could not achieve an accuracy higher than 0.401. Indeed, saving only facts relative to the question seems to be the key to achieving a good score in this task. We also noticed that using plain sentences, even with QDREN, we cannot achieve a higher accuracy. This might be due to the sentence encoder, since just using the multiplicative masks does not provide enough expressive power for getting key features of the sentence. Moreover, we notice that the accuracy achieved in the training set is always lower than that in the validation and test set. The same phenomenon is present also in other models, in our particular case this might be due to the strong regularization term used in our models. 
\begin{table}[t]
\centering
\caption{Validation/Test accuracy (\%) on CNN dataset. In the list AR stands for \emph{Attentive Reader}, AS for \emph{Attentive Sum}, AoA for \emph{Attention over Attention}, and DER for \emph{Dynamic Entity Representation}. \\}
\label{CNNris}
\scalebox{0.8}{
\begin{tabular}{r|cc|r|cc|r|cc}
\multicolumn{1}{l|}{}   & \textit{Val} & \textit{Test} & \multicolumn{1}{l|}{}     & \textit{Val} & \textit{Test} & \multicolumn{1}{l|}{} & \textit{Val} & \textit{Test} \\ \hline
\textbf{Max Freq.}      & 30.5         & 33.2          & \textbf{MemN2N}           & 63.4         & 66.8          & \textbf{AS Reader}    & 68.6         & 69.5          \\
\textbf{Frame-semantic} & 36.3         & 40.2          & \textbf{Attentive Reader} & 61.6         & 63            & \textbf{AoA}          & 73.1         &\textbf{74.4}          \\
\textbf{Word distance}  & 50.5         & 50.9          & \textbf{Impatient Reader} & 61.8         & 63.8          & \textbf{EpiReader}    & 73.4         & 74            \\
\textbf{LSTM Reader}    & 55           & 57            & \textbf{Stanford (AR)}    & 72.5         & 72.7          & \textbf{DER}          & 71.3         & 72.9          \\ \hline
\end{tabular}
}
\end{table}
Our model achieves an accuracy comparable to the \emph{Attentive and Impatient Reader} \cite{hermann2015teaching}, but not yet State-Of-The-Art model (i.e. \emph{Attention over Attention} (AoA)). It is worth noting though that our model is much simpler and it goes through each paragraph just once. A summary of the other models' results are shown in Table~\ref{CNNris}.  
\begin{figure}[b!]
\centering
  \begin{subfigure}[b]{\linewidth}
    \centering
        \caption{}
    \includegraphics[width=0.92\linewidth]{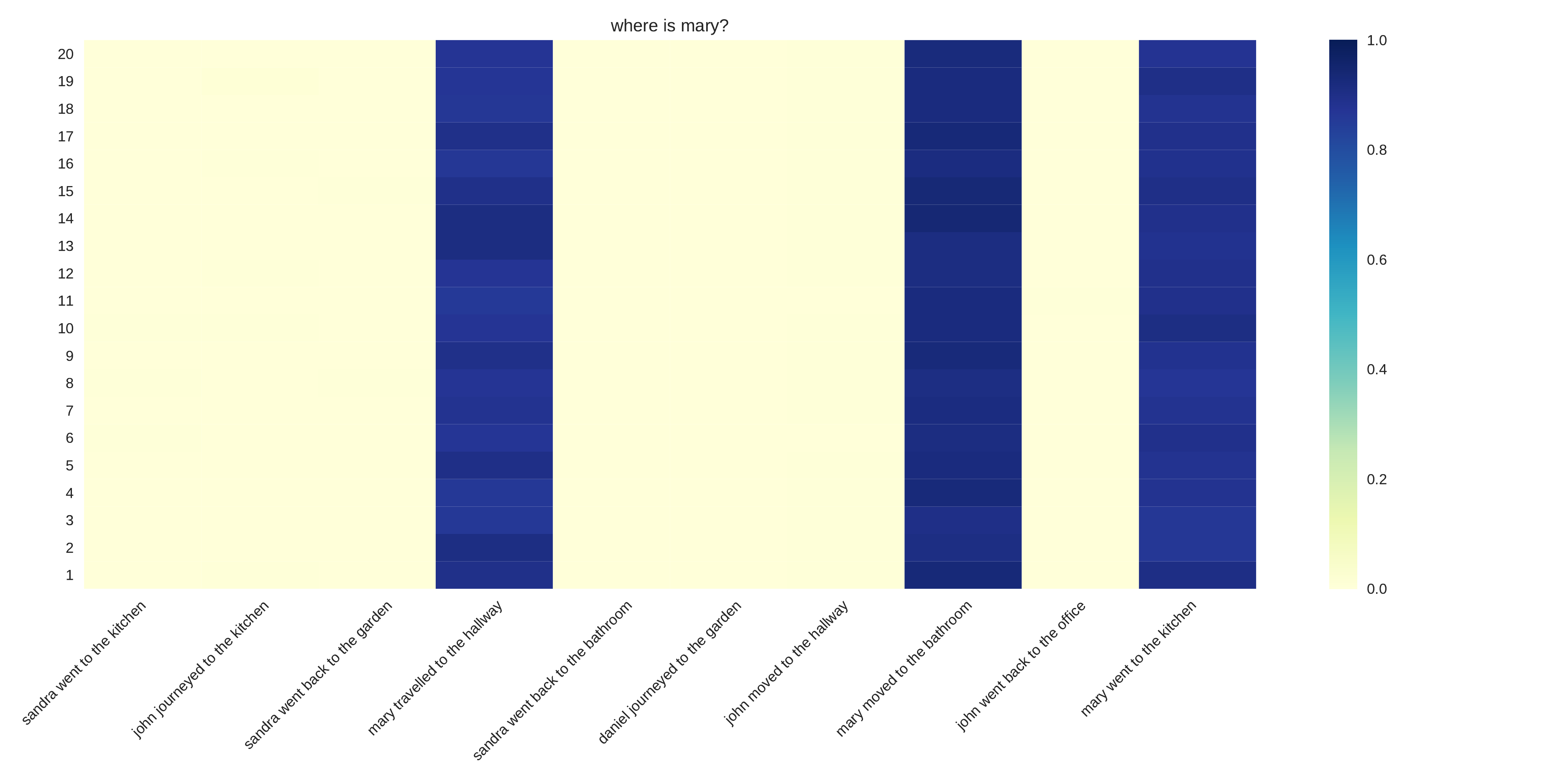}

  \end{subfigure}
\centering
  \begin{subfigure}[b]{\linewidth}
    \centering
        \caption{}
    \includegraphics[width=0.92\linewidth]{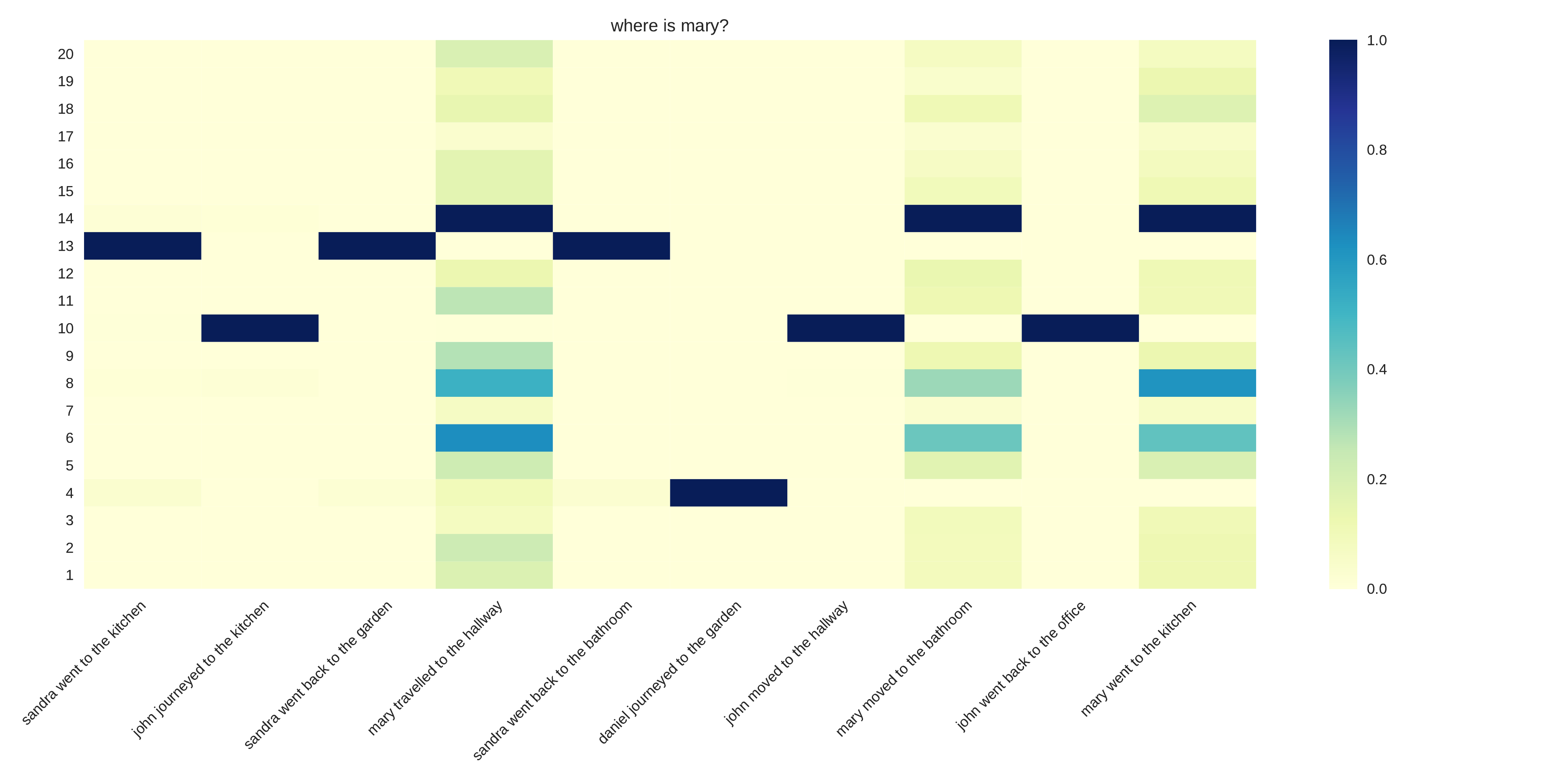}

  \end{subfigure}
\caption{Heatmap representing the gating function result for each memory block. In the y-axes represents the memory block number (20 in this example), in the x-axes, there are the sentences in the input divided into time steps, and at the top, there is the question to be answered. Darker color means a gate more open (values close to 1) and lighter colour means the gate less open. (a) shows QDREN and (b) shows REN.}
\label{Viz1}
\end{figure}

\section{Analysis}
To better understand how our proposed model (i.e. QDREN) works and how it improves the accuracy of the existing REN, we studied the gating function behavior. Indeed, the output of this function decides how much and what we store in each memory cell, and it is also where our model differs from the original one. Moreover, we trained the QDREN and the original REN using the bAbI task number 1 (using 20 memory blocks). We pick up this task since both models pass it, and it is one of the simplest, which also allows to better understand and visualize the results. Indeed, we have tried to visualize other tasks but the result was difficult to understand since there were too many sentences in input and it was difficult to understand how the gate opened. The visualization result is shown in Figure~\ref{Viz1}, where we plotted the activation matrix for both models, using a sample of the validation set. In these plots, we can notice how the two models learn which information to store. 
\\
\\
In Figure~\ref{Viz1} (a), we notice that the QDREN is opening the gates just when in the sentence appears the entity named Mary. This because such entity is also present in the question (i.e., "where is Mary?"). Even though the model is focusing on the right entity, its gates are opening all at the same times. In fact, we guess that a sparser activation would be better since it may have modeled which other entities are relevant for the final answer. Instead, the gaiting activation of the original REN is sparse, which is good if we would like to learn all the relevant facts in the text. Indeed, the model effectively assigns a block to each entity and it opens the gates just ones such entity appears in the input sentences. For example, in Figure~\ref{Viz1} (b) the block cell number 13 supposedly represents the entity Sandra, since each sentence in which this name appears the gate function of the block fully opens (value almost 1). Futher, we can notice the same phenomenon with the entity John (cell 10), Daniel (cell 4), and Mary (cell 14). Other entities (e.g., kitchen, bathroom, etc.) are more difficult to recognize in the plot since their activation is less strong and probably distributes this information among blocks.

\section{Conclusion}
In this paper we presented the \emph{Question Dependent Recurrent Entity Network}, used for reasoning and reading comprehension tasks. This model uses a particular RNN cell in order to store just relevant information about the given question. In this way, in combination with the original \emph{Recurrent Entity Network} (keys and memory), we improved the State-of-The-Art in the bAbI 1k task and achieved promising results in the Reading comprehension task on the \emph{CNN \& Daily news} dataset. However, we believe that there are still margins for improving the behavior for the proposed cell. Indeed, the cell has not enough expressive power to make a selective activation among different memory blocks (notice in Figure~\ref{Viz1} (a) the gates open for all the memories). This does not seem to be a serious problem since we actually outperform other models, but it could be the key to finally pass all the bAbI tasks. 

\section*{Acknowledgments}
This work has been supported in part by grant no. GA\_2016\_009 "Grandi Attrezzature 2016" by the University of Pisa.
%
%

\bibliographystyle{apalike}
\bibliography{biblio.bib}

\end{document}